\documentclass{IEEEtran}
\usepackage[latin9]{inputenc}
\usepackage{verbatim}
\usepackage{float}
\usepackage{wrapfig}
\usepackage{textcomp}
\usepackage{mathrsfs}
\usepackage{amsmath}
\usepackage{amsthm}
\usepackage{amssymb}
\usepackage{graphicx}

\makeatletter

\providecommand{\tabularnewline}{\\}

\theoremstyle{plain}
\newtheorem{thm}{\protect\theoremname}

\makeatother

\providecommand{\theoremname}{Theorem}

\begin{document}
\title{Learning non-rigid surface reconstruction from spatio-temporal image
patches}
\author{Matteo Pedone, Abdelrahman Mostafa and Janne Heikkilä\\
\emph{Center for Machine Vision Research and Signal Analysis, University
of Oulu, Finland}\\
\texttt{\{matteo.pedone, abdelrahman.mostafa, janne.heikkila\}@oulu.fi}}
\maketitle
\begin{abstract}
We present a method to reconstruct a dense spatio-temporal depth map
of a non-rigidly deformable object directly from a video sequence.
The estimation of depth is performed locally on spatio-temporal patches
of the video, and then the full depth video of the entire shape is
recovered by combining them together. Since the geometric complexity
of a local spatio-temporal patch of a deforming non-rigid object is
often simple enough to be faithfully represented with a parametric
model, we artificially generate a database of small deforming rectangular
meshes rendered with different material properties and light conditions,
along with their corresponding depth videos, and use such data to
train a convolutional neural network. We tested our method on both
synthetic and Kinect data and experimentally observed that the reconstruction
error is significantly lower than the one obtained using other approaches
like conventional non-rigid structure from motion.

\end{abstract}

\global\long\def\fun#1#2#3{#1:#2\rightarrow#3}%
 
\global\long\def\rn#1{\mathbb{R}^{#1}}%
\global\long\def\setdef#1#2#3{#1\coloneqq\left\{  #2:#3\right\}  }%
\global\long\def\sheargroup{A}%
\global\long\def\stretchgroup{S}%
\global\long\def\transgroup{T}%
\global\long\def\gauss#1{\mathscr{G}_{\mathrm{#1}}}%
\global\long\def\mat#1{\mathrm{\mathbf{#1}}}%
\global\long\def\databasesize{153600}%
\global\long\def\nframes{16}%
\global\long\def\patchsize{64\times64}%
\global\long\def\depthsize{32\times32}%

\section{Introduction}

\begin{wrapfigure}{r}{0.5\columnwidth}%
\begin{centering}
\includegraphics[width=3cm]{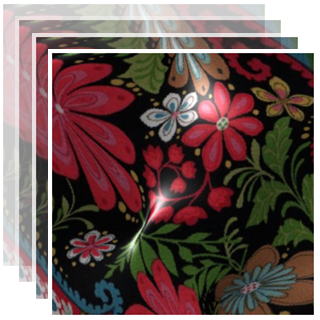} \includegraphics[height=4cm]{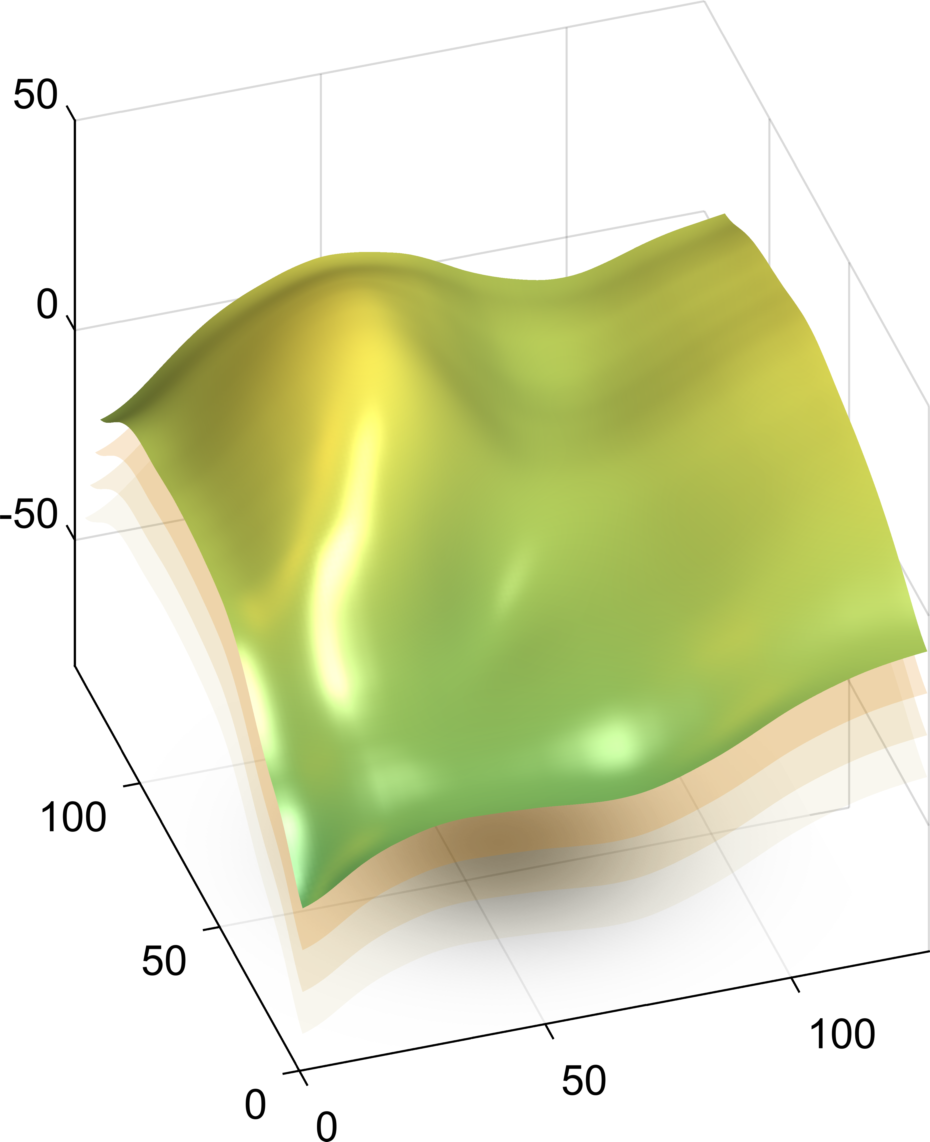}
\par\end{centering}
\caption{\label{fig:teaser}Frames of a $256\times256\times16$ video of a
non-rigidly deforming object and the corresponding depth maps obtained
directly from the video after training a CNN with a synthetically
generated database.}

\end{wrapfigure}%
The human visual system has a remarkable ability of discerning the
three-dimensional shape of an observed scene. Although its internal
mechanisms are still not entirely clear, it is known that our visual
system relies simultaneously on several cues to perceive shape \cite{shapiro_computer_2001},
such as the local changes in shading and texture of an object, or
the temporal change in appearance of an object while seen from different
angles. Researchers in the last few decades have attempted with varying
degrees of success to formulate mathematical models that would approximately
describe such mechanisms in order to emulate them on machines. Some
of the most prominent outcomes of these efforts are: algorithms for
\emph{structure from motion }(S\emph{f}M)\emph{ \cite{ullman_interpretation_1979},
photometric stereo} (PS) \cite{woodham_photometric_1978}, \emph{shape
from shading} \cite{ikeuchi_numerical_1981}, and \emph{shape from
texture} \cite{aloimonos_shape_1988}, where the last two mainly focus
on the special case of inferring shape from a single image. Each of
these methods has a vast literature of its own. Interesting surveys
can be found found in \cite{ozyesil_survey_2017,ackermann_survey_2015,ruo_zhang_shape--shading_1999}.
All the aforementioned methods have been typically approached under
the strong assumption that the observed scene is static. In recent
years, we have witnessed an increasing interest in extending such
techniques to the case of dynamic scenes made of rigidly and non-rigidly
moving objects. However, there are currently still many challenges
to be faced due to the ill-posedness of the inverse problems related
to the estimation of three-dimensional shape of a dynamic scene from
two-dimensional images. In photometric stereo, one or more images
of the same subject need to be acquired under different illuminations
in order to estimate its 3D shape, and some authors have addressed
the problem of moving objects by using specialized hardware \cite{fyffe_single-shot_2011,kim_photometric_2010,vogiatzis_practical_2010}.
In addition to that, the physical surface of the observed object is
assumed to have certain reflectance properties (usually Lambertian).
Similarly, in S\emph{f}M it is assumed that the spatial coordinates
in the image plane of some interest points are acquired from several
point of views, and under the assumption of a static scene, the geometric
relationships between corresponding points can be utilized to recover
the original 3D structure of the scene. The more general problem of
non-rigid structure from motion (NRS\emph{f}M) considers the case
where the scene itself can move non-rigidly, hence such geometric
relationships become more difficult to exploit. Researchers have introduced
priors on the camera model \cite{bregler_recovering_2000}, on the
type of trajectories of points in space-time \cite{akhter_trajectory_2011},
or on the shape of the object \cite{bregler_recovering_2000,kumar_multi-body_2016,dai_dense_2017,torresani_nonrigid_2008,yan_factorization-based_2008,kumar_scalable_2018}
in order to make the problem mathematically tractable. Algorithms
for NRS\emph{f}M mainly differ from each other in terms of the prior
they impose on the shape and motion of the objects, however currently,
most methods rely on the assumption that the scene is observed through
an orthographic camera (see Chapter 5 of \cite{salzmann_deformable_2010}
for some exceptions using the perspective camera model), and they
are essentially generalizations of the popular \emph{factorization
method}, which was originally employed for traditional orthographic
S\emph{f}M \cite{tomasi_shape_1992}. The same assumption of orthographic
camera model is made in most photometric stereo algorithms as well.
From a mathematical point of view, dynamic PS and NRS\emph{f}M aim
at solving different problems: the former relies on pixel intensities
from the images under different illuminations and returns normal maps
representing surfaces in 3D space, while the latter takes a set of
trajectories of tracked points in the image plane, and estimates their
corresponding representation in 3D space, as well as the trajectory
of the camera. Despite this fact, there exist some aspects of overlap
between these two methods which motivate our work.

Although both dynamic PS and NRS\emph{f}M differ significantly in
terms of the inputs they rely on and of the outputs they produce,
we can nonetheless argue that NRS\emph{f}M algorithms are often preceded
by tracking algorithms that in fact rely on information extracted
from the pixel values of a video sequence. Given the recent success
of deep-learning algorithms in many different settings, it is natural
to investigate the feasibility of a deep-learning based approach that
operates directly on a video sequence as its input, and that would
automatically yield a dense representation of the scene depth for
each frame. This has potentially the advantage of overcoming the frequently
imposed assumptions on the illumination conditions of the scene and
on the reflectance properties of the observed object (as in PS), and
that of avoiding to explicitly solve the factorization problem, as
in NRS\emph{f}M. 

The main challenge is clearly represented by the difficulty of acquiring
large scale training data in which each sample would consist of a
video sequence of an object (or a portion of it), along with its corresponding
3D representation in space. Although nowadays it is theoretically
possible to acquire RGB video sequences along with their corresponding
depth maps using inexpensive equipment, producing a sufficiently large
database involving non-rigid motion of many different subjects, materials,
textures, and light conditions would be a daunting and time consuming
process. 

In this manuscript, we try to address this problem by showing that
it is possible to generate an artificial database consisting of short
rendered video sequences, each depicting the movement of a small patch
of a non-rigidly deforming object, along with the corresponding depth
video, and then train a convolutional neural network to perform the
direct estimation of a depth video from the pixel intensities of a
video sequence.

The contributions of this paper are the following ones: first, we
propose a simple and computationally fast mathematical model to generate
the local spatio-temporal geometry of non-rigidly deforming surfaces,
which can be used to form a database of rendered video clips along
with their corresponding depths. Secondly, we propose a network architecture
to estimate the depth map for each frame of a video sequence of a
deforming object; thirdly, since the rendered scenes with an orthographic
camera model are often ambiguous with respect to stretches, shears
and translations along the optical axis, we derive a complete invariant
for this family of transformations that is used to formulate a loss
function in the training stage that implicitly operates on equivalence
classes of depth maps under such transformations.

The rest of this manuscript is organized as follows. Section \ref{sec:Our-Method}
describes the proposed method from a general point of view and introduces
the main assumptions that are invoked in the next sections. In Section
\ref{sec:Dataset-generation} we provide the details related to the
generation of our artificial database of spatio-temporal patches with
their corresponding ground-truth depth video. Sections \ref{sec:Invariants}-\ref{sec:Reconstruction}
contain the mathematical details related to the estimation of the
depth video, while the architecture of the convolutional neural network
(CNN) utilized for 3D shape estimation is described in Section \ref{sec:CNN-structure}.
Sections \ref{sec:Experiments} and \ref{sec:Conclusion-and-future}
respectively show the experimental results and give concluding remarks.

\section{Related work}

There are some recent works in the literature that can be considered
related to ours. Kumar et el. \cite{kumar_dense_2019} recently proposed
a method based on superpixels and motion constraints in order to produce
animated depth maps for a video sequence of moving subjects. Their
method does not rely on machine learning and the estimated depth maps
are always piecewise planar within the superpixel boundaries, affecting
the accuracy of the reconstruction. In \cite{brickwedde_mono-sf_2019}
a deep-learning based approach is presented to estimate the 3D scene
flow from a 2D video. However the main application domain is road
traffic navigation, and the scene is assumed to be approximately decomposable
into flat planes, while the motion of the objects is assumed to be
rigid. Agudo et al. shows in \cite{agudo_sequential_2016} that it
is feasible to solve the NRS\emph{f}M problem from video frames of
an orthographic monocular camera if one assumes that the non-rigid
motion can be accurately described by the equations governing the
motion of elastic bodies. After determining the 3D locations of the
tracked points, the authors propose a scheme to construct a triangular
mesh approximating the global shape of the surface. Despite the promising
results of existing methods, to the best of our knowledge, there are
still no approaches entirely based on deep-learning that attempt to
reconstruct the deforming surface of a non-rigidly moving object directly
from a sequence of images. 

\section{Overview of the proposed method\label{sec:Our-Method} }

We consider the problem of reconstructing the spatio-temporal depth
map (depth video) from the video sequence of a non-rigidly moving
object seen from a static orthographic camera. Our approach relies
on the following assumptions:
\begin{enumerate}
\item The scene/object is observed by a static orthographic camera.
\item The deformation of the object observable within fixed spatio-temporal
windows of the video sequence is non-negligible.
\item \label{enu:assumption_local_movement}Locally, the 4D structure of
the object (i.e. its 3D shape evolving in time) can be approximated
with a parametric model controlled by a relatively small amount of
parameters.
\end{enumerate}
Note that these assumptions, despite being restrictive, are analogous
to the assumptions made by current state-of-the-art NRSfM methods
\cite{kumar_scalable_2018-2,kumar_dense_2019}. The proposed method
essentially consists in estimating the depth video locally, within
local ``patches'' in the space-time domain, and then in reconstructing
the entire depth map of the scene by ``stitching'' together the
local depth videos. The estimation is done by training a neural network
to learn the animated depth map within small spatio-temporal patches
of a video that have size $\patchsize$ pixels $\times\nframes$ frames
in our implementation. Given the practical difficulty of acquiring
a large amount of real data consisting of small video clips of deforming
subjects, along with their corresponding depth videos, we generate
a database of synthetic data, i.e. deforming planar meshes rendered
with variable textures, reflectance properties, and light conditions,
and we obtain their respective depth videos through ray casting \cite{roth_ray_1982}.
The database is utilized to train a CNN to directly estimate $\patchsize\times\nframes$
depth videos from RGB video sequences of the same size. In the final
step, the depth videos of each patch are combined together in order
to reconstruct the entire depth video. The reconstruction scheme that
we utilize is inspired by the well-known method of signal reconstruction
using the constant overlap-add (COLA) decomposition. Note that the
assumption of orthographic camera removes the need to train the network
with different camera parameters, however it introduces also an ambiguity
problem where same video sequences may correspond to different depth
maps. This issue is addressed in detail in Section \ref{sec:Invariants}.

\section{Artificial dataset generation\label{sec:Dataset-generation}}

\begin{wrapfigure}{r}{0mm}%
\centering{}\includegraphics[width=4cm]{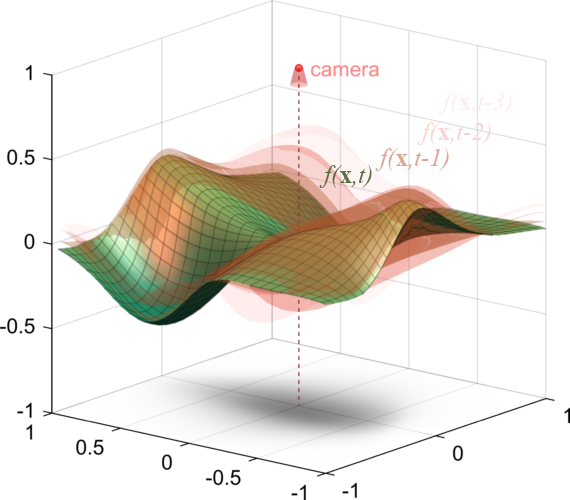}\caption{Graphical illustration of a spatio-temporal patch $f$ in \eqref{eq:vdm} }
\end{wrapfigure}
\begin{figure}[h]
\begin{centering}
\includegraphics[width=2.5cm]{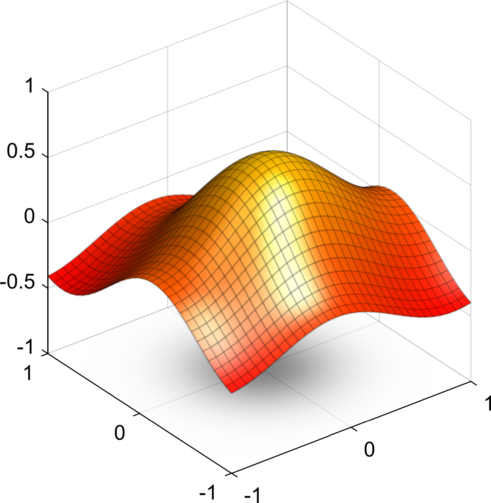} \includegraphics[width=2.5cm]{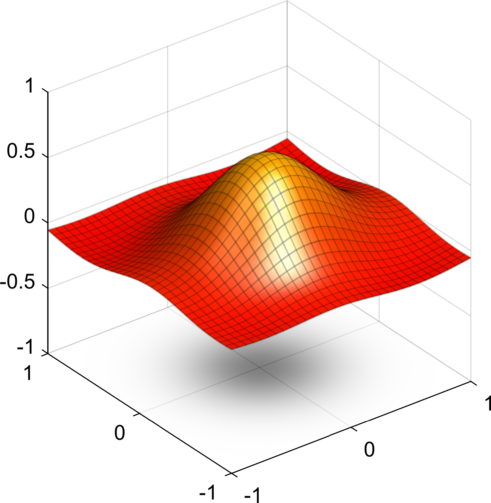}
\includegraphics[width=2.5cm]{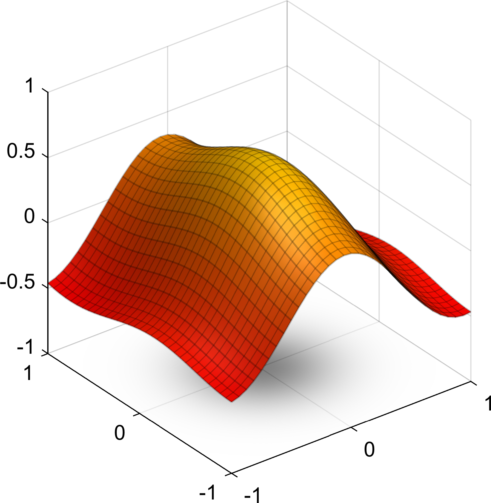} \includegraphics[width=2.5cm]{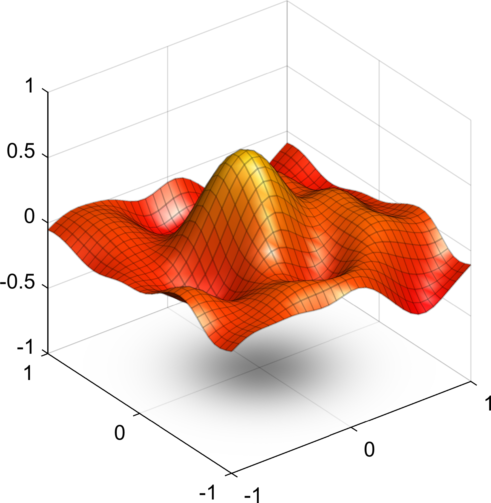}
\includegraphics[width=2.5cm]{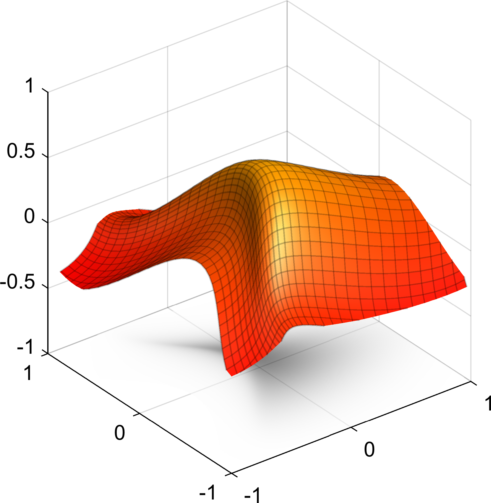}
\par\end{centering}
\caption{\label{fig:mesh_parameters}Illustration of the effect of the parameters
of \eqref{eq:vdm_Fourier}. From left to right, top to bottom: a patch
obtained by setting $\nu$, $\zeta$, $\xi$ to very low values and
$\Sigma=I$. The same patch with increased \emph{constraint}. With
non-identity $\Sigma$. With increased \emph{flexibility. }With increased
\emph{folding.} }
\end{figure}

\begin{figure}[H]
\begin{centering}
\includegraphics[width=2cm]{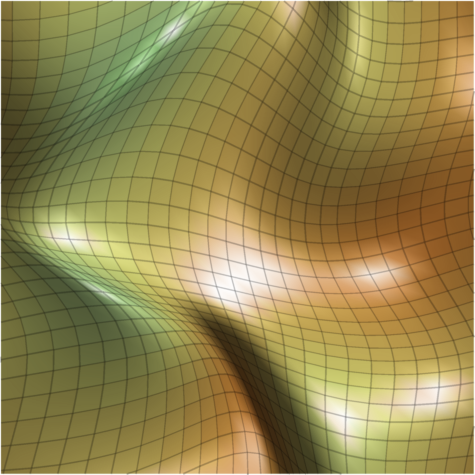} \includegraphics[width=2cm]{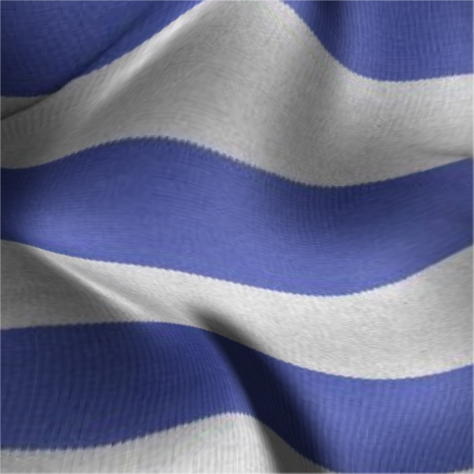}
\includegraphics[width=2cm]{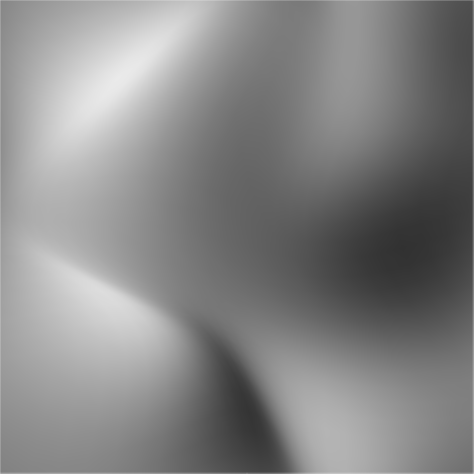}\\
\vspace{1mm}
\par\end{centering}
\begin{centering}
\includegraphics[width=2cm]{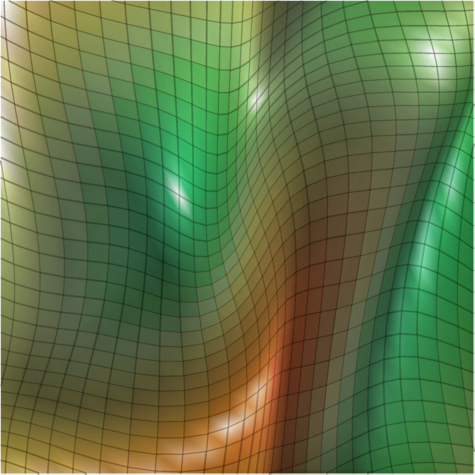} \includegraphics[width=2cm]{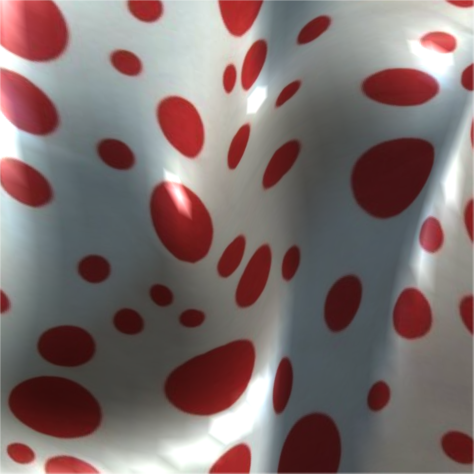}
\includegraphics[width=2cm]{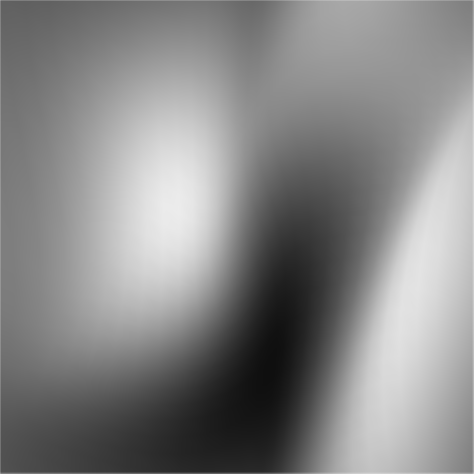}
\par\end{centering}
\caption{\label{fig:database_entry}Two examples of scenes rendered with our
method. From left to right, the 3D mesh, rendered object, and depth
map are depicted.}

\end{figure}

%
 Consider a smooth spatio-temporal image patch $\fun f{(-1,1)^{2}\times(0,1)}{\rn 3}$
having the following form:
\begin{equation}
f(\mathbf{x},t)=E_{t}\left(\mathbf{x}+d(\mathbf{x},t)\right)\label{eq:vdm}
\end{equation}
The quantity inside brackets in \eqref{eq:vdm} can be interpreted
as a 2D surface in $\rn 3$ obtained by applying a displacement to
each point of the domain of $f$. This essentially models a surface
obtained by applying a vector displacement (i.e. a displacement in
each of the three directions) to each point of the bi-unit square.
$E_{t}$ are Euclidean transformations (roto-translations) smoothly
parametrized by $t$ that change the orientation and position of the
patch. Let us denote $\gauss{\Sigma}(\mathbf{u})=e^{-\mathbf{u}^{T}\mathrm{\Sigma}\mathbf{u}}$
as the two-dimensional Gaussian function where $\mathrm{\Sigma}$
is a $2\times2$ symmetric positive definite matrix, and a vector-valued
function $\varphi(\mathbf{u},t)$ as: 

\begin{equation}
\varphi(\mathbf{u},t)=\left[\begin{array}{c}
\cos\left(t\phi_{1}(\mathbf{u})+\theta_{1}(\mathbf{u})\right)\\
\cos\left(t\phi_{2}(\mathbf{u})+\theta_{2}(\mathbf{u})\right)\\
\cos\left(t\phi_{3}(\mathbf{u})+\theta_{3}(\mathbf{u})\right)
\end{array}\right]\label{eq:phase}
\end{equation}

\begin{equation}
d(\mathbf{x},t)=\kappa\,\mathscr{G}_{\nu}(\mathbf{x})\,\mathrm{diag}(\zeta,\zeta,1)\,\mathcal{F}^{-1}\left\{ w\,\mathscr{G}_{\xi}\,\gauss{\Sigma}\,\varphi(\mathbf{\cdot},t)\right\} \left(\mathbf{x}\right)\label{eq:vdm_Fourier}
\end{equation}
where $\mathcal{F}^{-1}$ is the inverse 2D Fourier transform operator
applied ``channel-wise'', and $\fun w{\left(\mathbb{R}_{\geq0}\right)\,^{2}}{\left[0,1\right]}$
is a 2-dimensional circular symmetric function whose radial profile
is rapidly increasing, and such that $w(\boldsymbol{0})=0$, which
guarantees that the integral of $d$ is zero. The scalar parameters
$\kappa$ (\emph{intensity}), $\nu$ (\emph{constraint}), $\zeta$
(\emph{folding}), $\xi$ (\emph{flexibility})\emph{, }respectively
control the overall strength of the displacement, how much the square
patch remains anchored at its border, how much it tends to create
folds, and how much it behaves like a soft or hard surface. The matrix
$\mathrm{\Sigma}$ controls the distribution of \emph{orientations}
of the displacements, while $\fun{\phi,\theta}{\rn 2}{\rn 3}$ are
functions that determine the speed by which each frequency component
changes in time and their initial phase angle (Figure \ref{fig:mesh_parameters}).
Our approach of defining a moving surface by means of \eqref{eq:vdm_Fourier}
can be seen as a computationally cheap way to emulate the appearance
of a planar surface with given physical properties, animated by means
of realistic cloth simulation. We argue that in our application it
is not strictly necessary to obtain physical correctness of the movement,
thus we adopt the strategy of using colored noise to generate the
geometry of the patch and its movement. Analogous strategies are often
encountered in the literature of computer graphics, especially in
application related to realistic real-time rendering of fluids, e.g.
\cite{tessendorf_simulating_2001}. Note that in a practical scenario,
the domain of $f$ would be obviously discretized into a 3D array.
After the geometry of a moving patch is defined, we render $\databasesize$
video clips of moving patches $f$ by varying the parameters in \eqref{eq:vdm_Fourier},
the light source parameters, the texture of the surface, its reflectance
properties, and the variance of the noise that is added after the
render. For rendering, we adopt the Phong reflectance model. More
accurate reflectance models could be used to increase realism, at
the expenses of computational speed and parameters to handle. The
textures that we use in our implementation are $256\times256$ crops
extracted at random position from the images of the DTD database \cite{cimpoi_describing_2014}.
The render pass produces for each clip a stack of $\nframes$ images
of $\patchsize$ pixels that we call $f_{render}$. In addition to
the render pass, we use ray casting \cite{roth_ray_1982} to produce
another $\patchsize\times\nframes$ array $f_{depth}$ of corresponding
depth maps (Figure \ref{fig:database_entry}). A single database entry
is represented by a pair $(f_{render},\,f_{depth})$.

\section{Ambiguity-invariant representation of depth maps\label{sec:Invariants}}

Adopting an orthographic camera model has the advantage of avoiding
the need to train the network with different camera parameters, however
scenes rendered with orthographic projection are prone to ambiguities.
In the particular case of Lambertian surfaces illuminated by directional
light sources, the resulting ambiguity is the \emph{generalized bas-relief
ambiguity} (GBR) \cite{belhumeur_bas-relief_1997}, that has been
studied extensively in the context of \emph{shape from shading,} and
which is essentially a composition of shears and stretches along \emph{z,
}the optical axis\emph{. }Although the rendered surfaces in our database
are not always perfectly Lambertian, the reflectance model adopted
in the rendering stage is not sufficient to resolve entirely the ambiguity,
even in presence of texture and specular highlights. Therefore, any
rendered scene in our database is virtually indistinguishable from
all the other hypothetical rendered scenes undergoing an arbitrary
GBR transformation. This issue would surely impair the training stage,
hence it is crucial to find a representation of smooth depth map surface
that is invariant to the GBR transformation. Once, such an invariant
representation is obtained, it is possible to utilize it in the loss
function of the training stage of the network. This has the big advantage
of letting the network treat all the GBR-transformed versions of one
depth map as an entire equivalence class: in fact, the invariants
of two depth maps related by a GBR transformation will be exactly
the same, hence the loss function will yield always zero in such cases.

Let us consider the space $\mathcal{S}$ of smooth surfaces (spatial
patches) $\fun f{\rn 2}{\rn 3}$ defined as follows, and denote with
$\widetilde{f}$ a GBR-transformed version of $f$: 
\begin{equation}
\begin{array}{c}
f(x,y)=\left[\begin{array}{c}
x\\
y\\
z(x,y)
\end{array}\right]\\
\\
\widetilde{f}=\left[\begin{array}{c}
\widetilde{x}\\
\widetilde{y}\\
\widetilde{z}
\end{array}\right]=\left[\begin{array}{c}
x\\
y\\
\alpha x+\beta y+\lambda z+\tau
\end{array}\right]
\end{array}\label{eq:height map}
\end{equation}
where $z\in C^{\infty}(\rn 2)$ is the corresponding depth map, $\text{\ensuremath{\alpha,\beta,\tau\in\rn{}}}$,
and $\lambda\in\rn +$. In order to find a representation for an arbitrary
$f$ that is invariant to the group of GBR transformations, we use
the \emph{normalization construction} \cite{olver_classical_1999},
which is a well-known technique in classical invariant theory that
enables us to algorithmically derive complete differential invariants
for the action of the GBR transformations on depth maps. We thus make
the following statement (its proof can be found in the supplementary
material): 
\begin{thm}
\label{thm:GBR_invariant}Given a surface $f\in\mathcal{S}$, the
quantity 
\begin{equation}
\iota_{f}:=\frac{\nabla^{2}z}{\left\Vert \nabla^{2}z\right\Vert _{F}}\label{eq:GBR invariant}
\end{equation}
where $\nabla^{2}$ and $\left\Vert \cdot\right\Vert _{F}$ denote
respectively the Hessian and the Frobenius norm, is a complete differential
invariant for $f$ with respect to the action of the GBR transformations.
\end{thm}
In our application we consider also another complete invariant for
the subgroup of the GBR transformations given by scaling and translations
along $z$, and obtained from \eqref{eq:height map} by setting $\alpha=\beta=0$.
\begin{thm}
Given a surface $f\in\mathcal{S}$, the quantity 
\begin{equation}
\eta_{f}:=\left\Vert \nabla^{2}z\right\Vert _{F}^{-1}\left(\nabla z,\nabla^{2}z\right)\label{eq:BR invariant}
\end{equation}
where $\nabla,\nabla^{2}$ are respectively the gradient and Hessian
operators, is a complete differential invariant for $f$ with respect
to stretches and translations along $z$.
\end{thm}
Note that in a practical implementation, the partial derivatives of
$z$ in \eqref{eq:GBR invariant}-\eqref{eq:BR invariant} would need
to be estimated with first and second derivative filters.

\section{Network architecture\label{sec:CNN-structure}}

\begin{figure}[h]
\begin{centering}
\includegraphics[width=9cm]{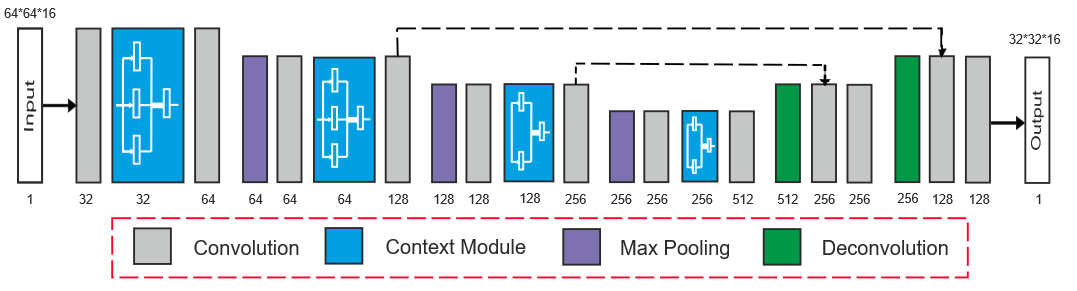}
\par\end{centering}
\caption{\label{fig:network}The network architecture: the input is a grayscale
video and the output is the corresponding sequence of depth maps.
Below each layer the number of channels is reported.}
\end{figure}

In order to estimate a depth video from its corresponding sequence
of grayscale video frames, we use an architecture similar to the 3D
U-net \cite{cicek_3d_2016} (see Figure \ref{fig:network}). The input
for the network is a $\nframes$ frames video and each frame size
is $\patchsize$ pixels. The output of the network is the corresponding
depth video, which has size $\depthsize\times\nframes$. To enlarge
the receptive field and get richer features, a context module \cite{zhou_davanet_2019}
with Atrous spatial pyramid pooling is introduced which performs parallel
dilated convolutions with different dilation rates. These feature
maps are then concatenated, and the output of the module is the convolution
of these feature maps. For the first two stages in the network, three
dilation rates are used ($1,2,3$). For the following two stages,
two dilation rates are used ($1,2$) as the feature maps dimensions
get smaller. We use $3\times3\times3$ kernels for all convolutions.
The \emph{leaky ReLU} activation function is used for all layers,
except for the last one, which has a \emph{linear} activation function
to predict the depths. 

\section{Reconstruction of a depth map from local patches\label{sec:Reconstruction}}

The proposed network architecture is designed to recover only depth
videos whose frames have dimension $\patchsize$. When the input video
has larger dimensions, an additional step is required in order to
reconstruct the full-sized depth map from the local patches. To address
this issue we split a frame of the input video into $\patchsize$
squares with constant overlap and run our algorithm to estimate the
depth in each block. Since the recovered depth maps are ambiguous
with respect to a GBR transformation, we also run our algorithm on
a version of the input video downsized along the spatial dimension
to $\patchsize$. This yields depth maps that represent a coarse representation
of the full-sized depth map. We upscale the coarse depth map to the
original size of the video frames, and to each $\patchsize$ patch
we apply the best (in least square sense) GBR transformation that
aligns it with the corresponding patch in the coarse level. The resulting
patches are then multiplied by a two-dimensional triangle function
(i.e. the outer product of a triangle function with itself) and added
together. Note that such a reconstruction scheme has strong analogies
with the well-known constant overlap-add (COLA) reconstruction often
seen in the context of reconstruction of signals from their short-time
Fourier transform \cite{allen_unified_1977}. The temporal dimension
is processed in an analogous way.

\section{Experiments\label{sec:Experiments}}

In order to validate our method we performed three experiments. In
the first one, we generated a database of $\databasesize$ entries
as explained in Section \ref{sec:Dataset-generation}. We split it
into three partitions: $80\%$ for training, $10\%$ for validation,
$10\%$ for testing, and use it to train a CNN with the architecture
described in Section \ref{sec:CNN-structure}. For training, we tried
two loss functions: $\ell_{GBR}=\mathrm{MSE}(\iota_{f},\iota_{f^{*}})$
and $\ell_{tr+sc}=\mathrm{MSE}(\eta_{f},\eta_{f^{*}})$, where $f,f^{*}$
are respectively the estimated and ground truth depth maps. Since
both invariants can be degenerate at some locations, we consider only
those pixels where the quantities are defined. To quantitatively evaluate
the performance we align the estimated depth maps with the corresponding
ground truth by a linear transformation and calculate the\emph{ mean
absolute spatially-normalized error} $\mathrm{MAE-SN}=T^{-1}\sum_{t=1}^{T}\sigma_{f^{*}(\cdot,t)}^{-1}\mathrm{MAE}(f(\cdot,t),f^{*}(\cdot,t))$,
which is scale-invariant in the spatial dimensions. We also repeat
the same experiment by re-using the parameters for the alignment of
the first frame for all the other frames. The results are summarized
in Table \ref{tab:synthetic_data}. 
\begin{table}
\begin{centering}
\begin{tabular}{|c|c|c|}
\cline{2-3} \cline{3-3} 
\multicolumn{1}{c|}{} & $\ell_{tr+sc}$ & $\ell_{GBR}$\tabularnewline
\hline 
{\scriptsize{}align each frame} & 0.4258 \textpm{} 0.1384 & \textbf{0.3832 \textpm{} 0.1305}\tabularnewline
\hline 
{\scriptsize{}alignment from 1st frame} & 0.6657 \textpm{} 0.2718 & \textbf{0.6133 \textpm{} 0.259}\tabularnewline
\hline 
\end{tabular}
\par\end{centering}
\caption{\label{tab:synthetic_data} Average and standard deviation of MAE-SN
calculated from 15360 video sequences for the two loss functions considered.}
\end{table}

In the second experiment we synthetically generate $1000$ samples
$(f_{render},\,f_{depth})$ having size $256\times256\times16$ using
the method in Section \ref{sec:Dataset-generation} and use significantly
different parameters in Equation \eqref{eq:vdm_Fourier} than the
ones used in the training stage. We then automatically select $1089$
points scattered uniformly on the 3D mesh and track their $xy-$coordinates
on the image plane. We use these coordinates as inputs for two popular
state-of-the-art NRSfM methods: CSF2 \cite{gotardo_non-rigid_2011}
, and KSTA \cite{gotardo_kernel_2011-1}, and obtain the \emph{z-}coordinates
of the tracked points for each frame. To obtain a dense $256\times256$
depth map we use scattered interpolation at each frame and then apply
a linear transformation to the estimated depth maps to align them
with the respective ground-truths and calculate the MAE-SN of each
depth video. We finally compare the results with the depth videos
obtained directly from the video sequences using our method. Since
the videos have larger dimensions than the ones used in training,
we apply the strategy described in Section \ref{sec:Reconstruction}
to obtain full-sized depth videos. The results are summarized in Table
\ref{tab:comparison_synthetic}.
\begin{table}
\begin{centering}
\begin{tabular}{|c|c|c|}
\hline 
ours with $\ell_{GBR}$ & CSF2 \cite{gotardo_non-rigid_2011} & KSTA \cite{gotardo_kernel_2011-1}\tabularnewline
\hline 
\hline 
\textbf{0.5907 \textpm{} 0.4536}  & 0.8746 \textpm{} 0.6372  & 0.8738 \textpm{} 0.6369 \tabularnewline
\hline 
\end{tabular}
\par\end{centering}
\caption{\label{tab:comparison_synthetic}Average and standard deviation of
MAE-SN calculated from 1000 video sequences using our method and two
state-of-the-art NRSfM reconstruction algorithms.}
\end{table}

\begin{figure}
\begin{centering}
\includegraphics[height=3.5cm]{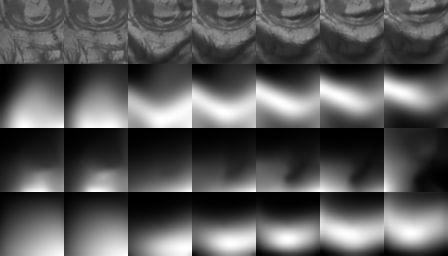}
\par\end{centering}
\caption{\label{fig:monkey_kinect1}\emph{First row:} seven frames extracted
from a video sequence depicting a detail of a shirt on a moving person.
\emph{Second row: }ground truth depth maps obtained with Kinect 1.
\emph{Third row: }depth maps recovered with KSTA \cite{gotardo_kernel_2011-1}
(CSF2 \cite{gotardo_non-rigid_2011} produced visually similar results)\emph{.
Last row: }depth maps recovered with our method. The contrast of the
depth maps in the figure was enhanced for visualization purposes.}
\end{figure}

In the third experiment we used real data, where $25$ frames of grayscale
video and the corresponding depth maps were acquired using Microsoft
Kinect 1 pointed to the shirt of a moving person in an indoor environment.
A fixed region of interest ($\patchsize$ pixels) was manually chosen,
where details were visible and depth pixels were defined, thus the
whole input video sequence has size $\patchsize\times25$.  For our
method, we estimate the depth videos directly from the video sequence,
as done in the previous experiments (Figure \ref{fig:monkey_kinect1}).
To obtain the input for CSF2 and KSTA algorithms, we performed point
tracking on the two video sequences using the minimum eigenvalue algorithm
\cite{jianbo_shi_good_1994}, which yielded the highest amount of
correctly tracked points (31)  and manually calibrated their parameters
to obtain best results. The depth videos were obtained as in the previous
experiment. The results are given in Table \ref{tab:comparison_real}.
\begin{table}
\begin{centering}
\begin{tabular}{|c|c|c|}
\hline 
ours with $\ell_{GBR}$ & CSF2 \cite{gotardo_kernel_2011} & KSTA \cite{gotardo_kernel_2011-1}\tabularnewline
\hline 
\hline 
\textbf{3.7} & 4.6 & 4.3\tabularnewline
\hline 
\end{tabular}
\par\end{centering}
\caption{\label{tab:comparison_real}Average and standard deviation of MAE
(in millimeters) calculated from a real video sequence using our method
and two state-of-the-art NRSfM reconstruction algorithms.}
\end{table}
 
\section{Conclusion \label{sec:Conclusion-and-future}}

We presented a deep-learning based approach to recover a depth video
directly from a video sequence of a non-rigidly deforming object.
We addressed the problem of training the network by synthetically
generating a large database of short videos depicting deforming 3D
meshes rendered with realistic textures, and paired with their corresponding
depth videos. We tested our method on both synthetic and real videos
and experimentally observed that the quality of the estimated depth
videos outperforms that of state-of-the-art NRSfM algorithms.

\section*{Appendix}

In this Appendix we provide a proof of Theorem \ref{thm:GBR_invariant}.
Since $\widetilde{x}=x$ and $\widetilde{y}=y$, we can concentrate
the discussion on $\widetilde{z}$. Consider the $n$-th order\emph{
prolongation }of $\widetilde{z}$ at an arbitrary point of $\rn 2$:
\begin{equation}
\widetilde{z^{[n]}}=\left[\begin{array}{ccccccc}
\widetilde{z} & \widetilde{z_{x}} & \widetilde{z}_{y} & \widetilde{z_{xx}} & \widetilde{z_{xy}} & \widetilde{z_{yy}} & \ldots\end{array}\right]\in\rn{[n]}\label{eq:prolongation}
\end{equation}
where the rightmost term is an element of a real vector space $\rn{[n]}$
having as many dimensions as the number of partial derivatives of
$z$ of order less than or equal to $n$. Consider the following system
of \emph{normalization equations}:
\begin{equation}
\left\{ \begin{array}{ccc}
\widetilde{z} & = & 0\\
\widetilde{z_{x}} & = & 0\\
\widetilde{z_{y}} & = & 0\\
\widetilde{z_{xx}}^{2}+\widetilde{z_{xy}}^{2}+\widetilde{z_{yx}}^{2}+\widetilde{z_{yy}}^{2} & = & 1
\end{array}\right.\label{eq:normalization equations}
\end{equation}
By plugging the explicit formulas for the partial derivatives in \eqref{eq:prolongation}
into \eqref{eq:normalization equations} and solving for the parameters
$\alpha,\beta,\lambda,\tau$ one obtains: 
\begin{equation}
\begin{array}{ccc}
\alpha & = & -\frac{z_{x}}{\left\Vert \nabla^{2}z\right\Vert _{F}}\\
\\
\beta & = & -\frac{z_{y}}{\left\Vert \nabla^{2}z\right\Vert _{F}}\\
\\
\lambda & = & \left\Vert \nabla^{2}z\right\Vert _{F}^{-1}\\
\\
\tau & = & -\left[\begin{array}{ccc}
\alpha & \beta & \lambda\end{array}\right]f
\end{array}\label{eq:moving frame}
\end{equation}
The equations in \eqref{eq:moving frame} are called \emph{moving
frame }and plugging the parameter values of the moving frame into
\eqref{eq:prolongation} yields a differential invariant for the action
of $G$: 
\begin{equation}
I_{f}=\frac{1}{\left\Vert \nabla^{2}z\right\Vert }\left[\begin{array}{cccccccc}
0 & 0 & 0 & z_{xx} & z_{xy} & z_{yx} & z_{yy} & \ldots\end{array}\right]\label{eq:invariantization}
\end{equation}
It is possible to verify, that the entries of $I_{f}$ that contain
derivatives of degree higher than 2 are obtainable from the firsts
three non-zero invariants in \eqref{eq:invariantization}. Furthermore,
we recognize $z_{xx}$, $z_{xy}$, $z_{yx}$, $z_{yy}$ as the entries
of the $2\times2$ Hessian matrix of $z$, therefore we can conclude
that a complete differential invariant for the action of $G$ on $\mathcal{S}$
is given by: 
\[
\iota_{f}=\frac{1}{\left\Vert \nabla^{2}z\right\Vert _{F}}\left[\begin{array}{cc}
z_{xx} & z_{xy}\\
z_{yx} & z_{yy}
\end{array}\right]=\frac{\nabla^{2}z}{\left\Vert \nabla^{2}z\right\Vert _{F}}
\]

\section*{Acknowledgements}

The authors would like to thank Prof. Peter J. Olver for his technical
advice and elucidations on the theory of the moving frame, and Academy
of Finland for the financial support for this research (grant no.
297732).

\bibliographystyle{plain}
\bibliography{cvpr2020}

\begin{thebibliography}{10}

\bibitem{ackermann_survey_2015}
Jens Ackermann and Michael Goesele.
\newblock A {{Survey}} of {{Photometric Stereo Techniques}}.
\newblock {\em Foundations and Trends\textregistered{} in Computer Graphics and
  Vision}, 9(3-4):149--254, November 2015.

\bibitem{agudo_sequential_2016}
Antonio Agudo, Francesc {Moreno-Noguer}, Bego{\~n}a Calvo, and J.~M.~M.
  Montiel.
\newblock Sequential {{Non}}-{{Rigid Structure}} from {{Motion Using Physical
  Priors}}.
\newblock {\em IEEE Transactions on Pattern Analysis and Machine Intelligence},
  38(5):979--994, May 2016.

\bibitem{akhter_trajectory_2011}
I.~Akhter, Y.~Sheikh, S.~Khan, and T.~Kanade.
\newblock Trajectory {{Space}}: {{A Dual Representation}} for {{Nonrigid
  Structure}} from {{Motion}}.
\newblock {\em IEEE Transactions on Pattern Analysis and Machine Intelligence},
  33(7):1442--1456, July 2011.

\bibitem{allen_unified_1977}
J.B. Allen and L.R. Rabiner.
\newblock A unified approach to short-time {{Fourier}} analysis and synthesis.
\newblock {\em Proceedings of the IEEE}, 65(11):1558--1564, November 1977.

\bibitem{aloimonos_shape_1988}
J.~Aloimonos.
\newblock Shape from texture.
\newblock {\em Biological Cybernetics}, 58(5):345--360, April 1988.

\bibitem{belhumeur_bas-relief_1997}
P.N. Belhumeur, D.J. Kriegman, and A.L. Yuille.
\newblock The bas-relief ambiguity.
\newblock In {\em Proceedings of {{IEEE Computer Society Conference}} on
  {{Computer Vision}} and {{Pattern Recognition}}}, pages 1060--1066, {San
  Juan, Puerto Rico}, 1997. {IEEE Comput. Soc}.

\bibitem{bregler_recovering_2000}
C.~Bregler, A.~Hertzmann, and H.~Biermann.
\newblock Recovering non-rigid {{3D}} shape from image streams.
\newblock In {\em Proceedings {{IEEE Conference}} on {{Computer Vision}} and
  {{Pattern Recognition}}. {{CVPR}} 2000 ({{Cat}}. {{No}}.{{PR00662}})},
  volume~2, pages 690--696 vol.2, June 2000.

\bibitem{brickwedde_mono-sf_2019}
Fabian Brickwedde, Steffen Abraham, and Rudolf Mester.
\newblock Mono-{{SF}}: {{Multi}}-{{View Geometry Meets Single}}-{{View Depth}}
  for {{Monocular Scene Flow Estimation}} of {{Dynamic Traffic Scenes}}.
\newblock {\em arXiv:1908.06316 [cs]}, August 2019.

\bibitem{cicek_3d_2016}
{\"O}zg{\"u}n {\c C}i{\c c}ek, Ahmed Abdulkadir, Soeren~S Lienkamp, Thomas
  Brox, and Olaf Ronneberger.
\newblock {{3D U}}-{{Net}}: Learning dense volumetric segmentation from sparse
  annotation.
\newblock In {\em International Conference on Medical Image Computing and
  Computer-Assisted Intervention}, pages 424--432. {Springer}, 2016.

\bibitem{cimpoi_describing_2014}
M.~Cimpoi, S.~Maji, I.~Kokkinos, S.~Mohamed, and {and}~A. Vedaldi.
\newblock Describing {{Textures}} in the {{Wild}}.
\newblock In {\em Proceedings of the {{IEEE Conf}}. on {{Computer Vision}} and
  {{Pattern Recognition}} ({{CVPR}})}, 2014.

\bibitem{dai_dense_2017}
Yuchao Dai, Huizhong Deng, and Mingyi He.
\newblock Dense {{Non}}-rigid {{Structure}}-from-{{Motion Made Easy}} - {{A
  Spatial}}-{{Temporal Smoothness}} based {{Solution}}.
\newblock {\em arXiv:1706.08629 [cs]}, June 2017.

\bibitem{fyffe_single-shot_2011}
Graham Fyffe, {Xueming Yu}, and Paul Debevec.
\newblock Single-shot photometric stereo by spectral multiplexing.
\newblock In {\em 2011 {{IEEE International Conference}} on {{Computational
  Photography}} ({{ICCP}})}, pages 1--6, April 2011.

\bibitem{gotardo_kernel_2011}
P.~F.~U. Gotardo and A.~M. Martinez.
\newblock Kernel non-rigid structure from motion.
\newblock In {\em 2011 {{International Conference}} on {{Computer Vision}}},
  pages 802--809, November 2011.

\bibitem{gotardo_kernel_2011-1}
Paulo F.~U. Gotardo and Aleix~M. Martinez.
\newblock Kernel non-rigid structure from motion.
\newblock In {\em 2011 {{International Conference}} on {{Computer Vision}}},
  pages 802--809, {Barcelona, Spain}, November 2011. {IEEE}.

\bibitem{gotardo_non-rigid_2011}
Paulo~F.U. Gotardo and Aleix~M. Martinez.
\newblock Non-rigid structure from motion with complementary rank-3 spaces.
\newblock In {\em {{CVPR}} 2011}, pages 3065--3072, June 2011.

\bibitem{ikeuchi_numerical_1981}
Katsushi Ikeuchi and Berthold K.~P. Horn.
\newblock Numerical shape from shading and occluding boundaries.
\newblock {\em Artificial Intelligence}, 17(1):141--184, August 1981.

\bibitem{kim_photometric_2010}
Hyeongwoo Kim, Bennett Wilburn, and Moshe {Ben-Ezra}.
\newblock Photometric {{Stereo}} for {{Dynamic Surface Orientations}}.
\newblock In Kostas Daniilidis, Petros Maragos, and Nikos Paragios, editors,
  {\em Computer {{Vision}} \textendash{} {{ECCV}} 2010}, Lecture {{Notes}} in
  {{Computer Science}}, pages 59--72, {Berlin, Heidelberg}, 2010. {Springer}.

\bibitem{kumar_multi-body_2016}
S.~Kumar, Y.~Dai, and H.~Li.
\newblock Multi-{{Body Non}}-{{Rigid Structure}}-from-{{Motion}}.
\newblock In {\em 2016 {{Fourth International Conference}} on {{3D Vision}}
  ({{3DV}})}, pages 148--156, October 2016.

\bibitem{kumar_scalable_2018}
Suryansh Kumar, Anoop Cherian, Yuchao Dai, and Hongdong Li.
\newblock Scalable {{Dense Non}}-rigid {{Structure}}-from-{{Motion}}: {{A
  Grassmannian Perspective}}.
\newblock {\em arXiv:1803.00233 [cs]}, March 2018.

\bibitem{kumar_scalable_2018-2}
Suryansh Kumar, Anoop Cherian, Yuchao Dai, and Hongdong Li.
\newblock Scalable {{Dense Non}}-{{Rigid Structure}}-{{From}}-{{Motion}}: {{A
  Grassmannian Perspective}}.
\newblock In {\em Proceedings of the {{IEEE Conference}} on {{Computer Vision}}
  and {{Pattern Recognition}}}, pages 254--263, 2018.

\bibitem{kumar_dense_2019}
Suryansh Kumar, Ram~Srivatsav Ghorakavi, Yuchao Dai, and Hongdong Li.
\newblock Dense {{Depth Estimation}} of a {{Complex Dynamic Scene}} without
  {{Explicit 3D Motion Estimation}}.
\newblock {\em arXiv:1902.03791 [cs]}, March 2019.

\bibitem{olver_classical_1999}
Peter~J Olver.
\newblock {\em Classical Invariant Theory}, volume~44.
\newblock {Cambridge University Press}, 1999.

\bibitem{ozyesil_survey_2017}
Onur Ozyesil, Vladislav Voroninski, Ronen Basri, and Amit Singer.
\newblock A survey of structure from motion.
\newblock {\em Acta Numerica}, 26:305--364, May 2017.

\bibitem{roth_ray_1982}
Scott~D Roth.
\newblock Ray casting for modeling solids.
\newblock {\em Computer Graphics and Image Processing}, 18(2):109--144,
  February 1982.

\bibitem{ruo_zhang_shape--shading_1999}
{Ruo Zhang}, {Ping-Sing Tsai}, J.E. Cryer, and M.~Shah.
\newblock Shape-from-shading: A survey.
\newblock {\em IEEE Transactions on Pattern Analysis and Machine Intelligence},
  21(8):690--706, August 1999.

\bibitem{salzmann_deformable_2010}
Mathieu Salzmann and Pascal Fua.
\newblock Deformable {{Surface 3D Reconstruction}} from {{Monocular Images}}.
\newblock {\em Synthesis Lectures on Computer Vision}, 2(1):1--113, December
  2010.

\bibitem{shapiro_computer_2001}
Linda~G. Shapiro and George~C. Stockman.
\newblock {\em Computer Vision}.
\newblock {Prentice Hall}, {Upper Saddle River, NJ}, 2001.

\bibitem{jianbo_shi_good_1994}
Jianbo Shi and Tomasi.
\newblock Good features to track.
\newblock In {\em 1994 {{Proceedings}} of {{IEEE Conference}} on {{Computer
  Vision}} and {{Pattern Recognition}}}, pages 593--600, June 1994.

\bibitem{tessendorf_simulating_2001}
Jerry Tessendorf et~al.
\newblock Simulating ocean water.
\newblock {\em Simulating nature: realistic and interactive techniques.
  SIGGRAPH}, 1(2):5, 2001.

\bibitem{tomasi_shape_1992}
Carlo Tomasi and Takeo Kanade.
\newblock Shape and motion from image streams under orthography: A
  factorization method.
\newblock {\em International Journal of Computer Vision}, 9(2):137--154,
  November 1992.

\bibitem{torresani_nonrigid_2008}
L.~Torresani, A.~Hertzmann, and C.~Bregler.
\newblock Nonrigid {{Structure}}-from-{{Motion}}: {{Estimating Shape}} and
  {{Motion}} with {{Hierarchical Priors}}.
\newblock {\em IEEE Transactions on Pattern Analysis and Machine Intelligence},
  30(5):878--892, May 2008.

\bibitem{ullman_interpretation_1979}
S.~Ullman and Sydney Brenner.
\newblock The interpretation of structure from motion.
\newblock {\em Proceedings of the Royal Society of London. Series B. Biological
  Sciences}, 203(1153):405--426, January 1979.

\bibitem{vogiatzis_practical_2010}
George Vogiatzis and Carlos Hern{\'a}ndez.
\newblock Practical {{3D Reconstruction Based}} on {{Photometric Stereo}}.
\newblock In Roberto Cipolla, Sebastiano Battiato, Giovanni~Maria Farinella,
  and Janusz Kacprzyk, editors, {\em Computer {{Vision}}}, volume 285, pages
  313--345. {Springer Berlin Heidelberg}, {Berlin, Heidelberg}, 2010.

\bibitem{woodham_photometric_1978}
Robert~J. Woodham.
\newblock Photometric {{Stereo}}.
\newblock Technical Report AI-M-479, {MASSACHUSETTS INST OF TECH CAMBRIDGE
  ARTIFICIAL INTELLIGENCE LAB}, June 1978.

\bibitem{yan_factorization-based_2008}
J.~Yan and M.~Pollefeys.
\newblock A {{Factorization}}-{{Based Approach}} for {{Articulated Nonrigid
  Shape}}, {{Motion}} and {{Kinematic Chain Recovery From Video}}.
\newblock {\em IEEE Transactions on Pattern Analysis and Machine Intelligence},
  30(5):865--877, May 2008.

\bibitem{zhou_davanet_2019}
Shangchen Zhou, Jiawei Zhang, Wangmeng Zuo, Haozhe Xie, Jinshan Pan, and
  Jimmy~S Ren.
\newblock {{DAVANet}}: {{Stereo Deblurring}} with {{View Aggregation}}.
\newblock In {\em Proceedings of the {{IEEE Conference}} on {{Computer Vision}}
  and {{Pattern Recognition}}}, pages 10996--11005, 2019.

\end{thebibliography}
 
\end{document}